\documentclass[conference]{IEEEtran}
\IEEEoverridecommandlockouts

\usepackage{amsmath,amssymb,amsfonts}
\usepackage{algorithmic}
\usepackage{graphicx}
\usepackage{textcomp}
\usepackage{xcolor}
\usepackage[a4paper, total={184mm,239mm}]{geometry}
\def\BibTeX{{\rm B\kern-.05em{\sc i\kern-.025em b}\kern-.08em
    T\kern-.1667em\lower.7ex\hbox{E}\kern-.125emX}}

\usepackage{tabularx}
\usepackage{booktabs}
\usepackage{url}
\usepackage{csquotes}
\usepackage{hyperref}
\usepackage{url}
\usepackage{listings}
\usepackage[capitalise]{cleveref}
\Crefname{figure}{Fig.}{Fig.}
\usepackage{algorithm}
\usepackage[small,compact]{titlesec} 
\titlespacing{\section}{2pt}{1pt}{1pt}
\titlespacing{\subsection}{2pt}{1pt}{1pt}
\titlespacing{\subsubsection}{3pt}{2pt}{2pt}
\usepackage{siunitx}
\newcolumntype{Y}{>{\centering\arraybackslash}X}

\usepackage[style=numeric,sorting=none]{biblatex}
\newcommand{\leaf}[1]{leaf L#1}
\begin{document}

\title{Late Breaking Results: Conversion of Neural Networks into Logic Flows for Edge Computing\vspace{-.4em}\thanks{
This work is funded by the European Union - European Research Council (ERC) Starting Grant - Project-ID 101219243. Views and opinions expressed are those of the author(s).
}}

\author{\IEEEauthorblockN{Daniel Stein\textsuperscript{1}, Shaoyi Huang\textsuperscript{2}, Rolf Drechsler\textsuperscript{3}, Bing Li\textsuperscript{4}, Grace Li Zhang\textsuperscript{1}}
\IEEEauthorblockA{\textsuperscript{1}TU Darmstadt, \textsuperscript{2}Stevens Institute of Technology, \textsuperscript{3}University of Bremen, \textsuperscript{4}TU Ilmenau\\
Email: \mbox{\{daniel.stein, grace.zhang}@tu-darmstadt.de\},} shuang59@stevens.edu, drechsler@uni-bremen.de, bing.li@tu-ilmenau.de
\vspace{-1.2em}
}

\maketitle

\begin{abstract}
Neural networks have been successfully applied in various resource-constrained edge devices, where usually central processing units (CPUs) instead of graphics processing units exist due to limited power availability. 
%In CPUs, often a huge number of dedicated hardware multipliers and adders are not available to process massive multiply-accumulate (MAC) operations in neural networks. Therefore, the execution efficiency of neural networks is compromised on CPUs, leading to high latency. 
%Although state-of-the-art techniques, e.g., compression of neural networks and efficient scheduling, are effective in many scenarios, they still 
State-of-the-art research still 
focuses on efficiently executing enormous numbers of multiply-accumulate (MAC) operations. 
However, CPUs themselves are not good at executing such mathematical operations on a large scale, since they are more suited to execute control flow logic, i.e., computer algorithms. 
To enhance the computation efficiency of neural networks on CPUs, in this paper, we propose to convert them into logic flows for execution. Specifically, neural networks are first converted into equivalent decision trees, from which decision paths with constant leaves are then selected and compressed into logic flows.
%by converting them into equivalent decision trees, which are further abstracted into logic flows.
%, which can take advantage of the characteristics of CPUs. 
%Specifically, neural networks are first converted into equivalent decision trees, which are further abstracted into logic flows.
Such logic flows consist of \textit{if} and \textit{else} structures and a reduced number of MAC operations. %, %instead of massive MAC operations in the original neural networks, 
%so that the computation efficiency can be enhanced. 
Experimental results demonstrate that the latency can be reduced by up to 14.9\,\% on a simulated RISC-V CPU without any accuracy degradation. - The code is open source at \url{https://github.com/TUDa-HWAI/NN2Logic}
\vspace{-1em}
\end{abstract}

\begin{IEEEkeywords}
Neural network on CPUs; Logic flows of neural networks; Edge computing\vspace{-.4em}
\end{IEEEkeywords}

\section{Introduction}
%The last decade has witnessed the successful application of neural networks on edge devices. 
Neural networks executed on edge devices can provide real-time AI processing closer to the data source. %, which reduces the dependence on the cloud. 
In neural networks, a huge number of multiply-accumulate (MAC) operations need to be executed. However, edge devices usually have general-purpose central processing units (CPUs) and do not have graphics processing units (GPUs) with parallel MAC circuits due to their power limitations.
CPUs are not good at executing such mathematical operations on a large scale.
The reason is that a CPU normally contains a large circuit block for logic execution, while only a limited number of dedicated multipliers and adders are available for executing MAC operations.
Therefore, the execution efficiency of neural networks is compromised on CPUs, leading to high latency.

Various methods have been proposed to enhance the execution efficiency of neural networks on CPUs. At algorithm level, pruning~\cite{han2024dtmm,10546870,10247868}, quantization~\cite{zhuo2022empirical,10137171}, 
knowledge distillation~\cite{MOSLEMI2024100605}, 
neural architecture search~\cite{lin2020mcunet},  
and dynamic decisions~\cite{10492471,10595861} have been applied to reduce the number of MAC operations. %Another direction to enhance the computation efficiency focuses on hardware optimizations. For example, 
At hardware level, 
state-of-the-art AI compilers~\cite{apachetvm,iree,torch,mlir} exploit hardware mapping, data reuse, memory allocation, and fetching~\cite{lin2021memory}, instruction scheduling to avoid bottlenecks~\cite{10745806}, optimization for memory latency hiding, loop-oriented  optimizations, and parallelization. %Furthermore, layer and operator fusion are also exploited to reduce memory accesses. 

The techniques described above are effective in many scenarios. However, they still focus on executing many MAC operations instead of examining the logic expression of neural networks.
%In fact,  not all CPUs have a large array of MAC units to execute the scheduled MAC operations and CPUs themselves are not good at executing such mathematical operations on a large scale, since they are more suitable to execute control flow logic. 
%Accordingly, the execution efficiency of neural networks is compromised on CPUs.
Even for heterogeneous CPU platforms extended with a neural processing unit, it is still crucial to examine the execution of neural networks on CPUs in a logic form to take advantage of all the available computation resources.
This study is still missing in the state of the art to date.

%Different from previous work, 
In this paper, we explore converting the execution of neural networks into logic flows%, similar to computer algorithms, 
to enhance the execution efficiency of neural networks on CPUs. Such logic flows consist of \textit{if} and \textit{else} structures and a reduced number of MAC operations, so that the computation efficiency on CPUs can be enhanced. 

\section{Methodology}
\label{sec:motivation}
%Contrary to the state-of-the-art work on directly compiling MAC operations of neural networks, we aim to convert neural networks into logic flows for efficient CPU execution. 
%Specifically, 
To convert a neural network into logic flows for efficient CPU execution, 
we first convert it to an equivalent decision tree with the method in~\cite{Aytekin2022NeuralNA}. To reduce the complexity of the decision tree, only training data is used for the tree construction, as described in Section \ref{sec:conversion}. 
Afterwards, we select those decision paths, traversing from the root of the decision tree to leaf nodes with constant classification results.
We then convert the execution of such paths into logic flows, as described in Section \ref{sec:selection}.  %, since executing logic flows requires fewer MAC operations than executing the original neural network. 
The remaining decision paths will be executed in the original format of the neural network, leading to a hybrid execution on CPUs in Section \ref{sec:hybrid}.

\begin{figure*}[ht]
\centering
\includegraphics[width=\textwidth]{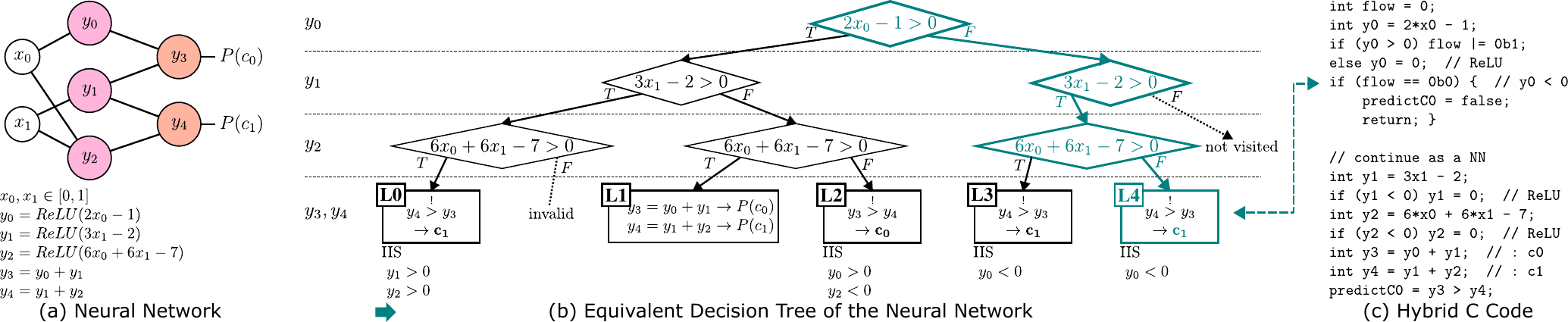}
\vspace{-1.1em}
  \caption{\centering{}Conversion of a neural network into an equivalent decision tree. (a)~A neural network for binary classification (classes $c_0$ and $c_1$). (b)~The decision tree of the neural network in (a) constructed using training data. (c)~C Code for hybrid execution of the neural network and the logic flow of \leaf{4}.}% \textcolor{red}{codes are too small to see. Change this to describe the leaf 4, and use arrow to show you are describing this corresponding path}}
\label{dt-example}
\end{figure*}

\Cref{dt-example} illustrates the concept of conversion, where a fully connected neural network in \cref{dt-example}(a) is converted into its equivalent decision tree in \cref{dt-example}(b). We use the extraction of the logic flow of the decision path to the right-most leaf, denoted as \leaf{4}, as an example. This leaf has a constant classification result of class $c_1$. An excerpt of C code for such a hybrid execution is shown in \cref{dt-example}(c). 
%\Cref{dt-example} illustrates this concept where a neural network in \cref{dt-example}(a) is converted into its equivalent decision tree in \cref{dt-example}(b) with the method described in~\cite{Aytekin2022NeuralNA}.

\begin{table*}[tp]
\caption{The runtime results of our hybrid execution.}% \textcolor{red}{Make this table smaller by removing Reference (cycles) and Hybrid (Cycles). Reviewers care about the reduction.}\textcolor{blue}{we do have a height problem not a width problem. even if I remove columns the width of the table is still too much for single column. the height of the table won't change.}}
\label{tab:exp_latency}

\setlength{\tabcolsep}{4pt}
\begin{tabularx}{\textwidth}{ccc *{2}{cYc} cYcc c}
    \toprule
    &&& \multicolumn{3}{c}{Reference (Cycles)} & \multicolumn{3}{c}{Hybrid (Cycles)} & \multicolumn{4}{c}{Latency Comparison} & \\
    \cmidrule(l{\tabcolsep}r{\tabcolsep}){4-6}
    \cmidrule(l{\tabcolsep}r{\tabcolsep}){7-9}
    \cmidrule(l{\tabcolsep}r{\tabcolsep}){10-13}
    Dataset & Network & Acc &
    Min & Avg &  Max &
    Min & Avg &  Max &
    Min & Avg & Max &
    Layer 2 + 3 & Exit by Tree \\
    \cmidrule(r{\tabcolsep}){1-3}
    \cmidrule(l{\tabcolsep}r{\tabcolsep}){4-6}
    \cmidrule(l{\tabcolsep}r{\tabcolsep}){7-9}
    \cmidrule(l{\tabcolsep}r{\tabcolsep}){10-13}
    \cmidrule(l{\tabcolsep}){14-14}

    MNIST\textsuperscript{*} & 25 - 30 - 2 & \num{0.937} & 
    \num{40326} & \num{40367.08} & \num{40436} &
    \num{38387} & \num{40095.32} & \num{40496} &
    \SI{4.8}{\percent} & \SI{0.7}{\percent} & \SI{-0.1}{\percent} & 
    \SI{19.1}{\percent} & \SI{15.11}{\percent} \\
    
    Occupancy I & 10 - 25 - 2 & \num{0.862} &
    \num{1650} & \num{1665.07} & \num{1691} &
    \num{789} & \num{1416.63} & \num{1712} &
    \SI{52.2}{\percent} & \SI{14.9}{\percent} & \SI{-1.2}{\percent} &
    \SI{39.3}{\percent} & \SI{34.1}{\percent} \\

    Occupancy II & 10 - 20 - 2 & \num{0.983} &
    \num{1292} & \num{1319.98} & \num{1329} &
    \num{963} & \num{1250.74} & \num{1337} &
    \SI{25.5}{\percent} & \SI{5.2}{\percent} & \SI{-0.6}{\percent} &
    \SI{25.1}{\percent} & \SI{12.7}{\percent} \\
    \bottomrule
\end{tabularx}
\label{tab:first-table}
\setlength{\tabcolsep}{6pt}
\end{table*}

%\subsection{Decision Tree Compression}

%\subsection{Logic Flows of the Neural Network}

%\section{Methodology}
%\label{sec:method}

%\textcolor{red}{If you want to chaneg something, please discuss with me first. }

\subsection{Conversion into Decision Trees with Training Data}
\label{sec:conversion}

%A prerequisite for this approach is that the activation functions need to be piecewise linear; in our case, the popular rectified linear unit (ReLU).
To construct the decision tree of a neural network, we build the decision nodes corresponding to the decision of the activation functions. 
Decision nodes are then connected via branches until we reach the last layer of the neural network.
For example, the input condition $x_0$ making $y_0$ larger than $0$ with ReLU as the activation function, i.e., $2x_0-1>0$, is set as the root node of the tree.
The input condition making $y_1$ larger than $0$, i.e., $3x_1-2>0$, is then constructed as decision nodes, which are connected with the root node via two branches representing the true~($T$) and false~($F$) decisions of the root node.
The remaining neurons can be processed similarly to create more decision nodes. 
We only add branches that are traversed by samples in the training dataset (e.g., the false branch of the right-most decision of $y_1$ in \cref{dt-example}(b) is never visited and thus not added).
When processing the neurons of the last layer, %the output of the neural network can be determined according to previous decision nodes. 
%Instead of creating decision nodes, 
we create leaf nodes, which are the final nodes of the decision tree. % after which, decision tree will not split the data. 
%E.g., in \cref{dt-example}(b) we determine all the activation regions of the ReLU ($f(x) > 0$ true or $f(x) \leq 0$ false) for each sample, and connect the nodes in a binary decision tree accordingly. 
%Finally, the last layer of the network is inserted as leaves at the end of each traversed branch.
%The constructed equivalent decision tree is showcased in \cref{dt-example}(b).

\subsection{Selection of Decision Paths for Extracting Logic Flows}
\label{sec:selection}
%Although the complexity of the constructed decision tree is reduced by using training data, the resulting decision tree can still be huge.
%In this step, we narrow down the potential branches of the decision tree to extract logic flows from.

%\subsubsection{Identification of Decision Paths with Constant Leaves}
After the construction of the decision tree, we will select those decision paths with constant leaves for logic flow extraction. 
A constant leaf in the case of a classification task means that one of the classes always appears, and the remaining classes never appear. For example, in \cref{dt-example}, the leftmost leaf, denoted as \leaf{0}, has the constant classification result of class $c_1$, indicating that all the training samples lead to the same classification.
To determine whether a decision path leads to a constant leaf for every possible input, we formulate a Mixed Integer Programming problem for every class on this path and try to solve it with Gurobi~\cite{gurobi}.
For example, for \leaf{4} we model two MIP problems, one for the case $y_4 > y_3$ (class $c_1$) and the other for $y_3 > y_4$ (class $c_0$). As there is no feasible solution for the case $y_3 > y_4$, this leaf will always yield $y_4$ larger than $y_3$, which means it always predicts class $c_1$.
%If the problem for a class is infeasible, it means the class will never be predicted by that leaf on that path.
Afterwards, we use Gurobi to determine 
the Irreducible Infeasible Subsystem (IIS) of constant leaves, which is a set of constraints that cause the problem to be infeasible, and if any of the constraints would be removed, the problem becomes solvable~\cite{Gleeson1990}.
For the example of \leaf{4} in \cref{dt-example}, the IIS is $y_0 < 0$, meaning class $c_1$ is always predicted because of the constraint $y_0 < 0$. We then use only this constraint to describe the logic flow of this decision path, since the other decision nodes are not required.
%All decision nodes not within the IIS can safely be removed from the decision path without affecting the constness of the leaf, thus simplifying the logic flow of the decision path. 

%\subsubsection{Compressing Decision Paths with Constant Leaves} 

%\subsubsection{Selection of Decision Paths for Latency Reduction}
%\label{sub:path_selection}

\subsection{Hybrid Execution with Selected Decision Paths}
\label{sec:hybrid}

A selected number of compressed paths with constant leaves are converted into logic flows.
For the remaining paths, they are executed in the original format of the neural network, leading to a hybrid execution. \Cref{dt-example}(c) shows the C code for the hybrid execution when only the decision path leading to \leaf{4} is converted into a logic flow.
All the neurons corresponding to constraints in the IIS of \leaf{4} are computed first. 
After all the required ReLU decisions are made, a tracking variable is checked to see if all the conditions of the logic flow are matched.
If this is the case, the output is determined, and the computation has finished.
After all logic flows have been processed, the remaining code is added, which continues computation as a regular neural network.

\section{Experimental Results}\label{sec:results}
\label{sec:experimental_results}

To verify the effectiveness of our approach, we applied it to three quantized 3-layer fully connected neural networks with three publicly available datasets: MNIST\footnotemark~\cite{mnist} and two room occupancy detection datasets, denoted as Occupancy I~\cite{occupancy} and Occupancy II~\cite{occupancy2}, as shown in \cref{tab:first-table}.
Since we generate bare-metal C code in our framework, we need a comparable reference implementation of the neural network for evaluation without the overhead of an interpreter.
Therefore, we also generate the Reference code, which has the same optimizations and ordering of neurons.
All implementations are evaluated on the hardware simulator of the Ibex RISC-V CPU~\cite{ibex}. 
%All neural networks consist of three layers, the structures of which are illustrated in the second column in Table~\ref{tab:first-table}, and have been quantized.  
%Since our framework currently supports binary classification, we modified MNIST to determine if a given training sample is either even or odd, denoted as MNIST\textsuperscript{*}. 
%All the neural networks are quantized. %The neural network processing MNIST\textsuperscript{*} is also pruned. The remaining neural networks are not pruned since they do not have much redundancy.
%The resulting neural networks are our starting point.
\footnotetext[1]{MNIST is modified to determine if a given training sample is either even or odd, denoted as MNIST\textsuperscript{*}.} %The relatively low accuracy (see \cref{tab:exp_latency}) stems from the quantization steps between layers, to keep the activations within \SI{8}{\bit}.}

%The Reference C code of a neural network is generated by describing the function of all the neurons in the neural network in C programming language. 
%This Reference code has the same optimizations and ordering of neurons as in the Hybrid code. 
%The Hybrid C Code generated by \textit{NN2Logic} and the Reference C Code of the neural networks are directly compiled for CPUs. 

%Compilation of the code for the CPU was performed using the RISC-V GNU Compiler Toolchain~\cite{rv-gcc}.
%Compiler optimizations were set to \enquote{size} (\verb|-Os|) since they yielded the lowest latency numbers for the Ibex CPU as well. The latency of each network is measured in cycles required for each inference sample in the inference dataset on the Ibex CPU hardware simulator. 

%\subsection{Latency Results} 
The results are shown in \cref{tab:first-table}. 
None of the three neural networks exhibits an accuracy degradation for the hybrid code compared with the reference (the model accuracies are shown in column \enquote{Acc}). 
%Latency results for the reference and the hybrid execution are expressed in clock cycles of the CPU.
%\enquote{Min} and \enquote{Max} mean  the minimum latency and the maximum latency of a sample in the inference dataset executing on the RISC-V CPU, respectively. \enquote{Avg} indicates the average latency over all the samples.
%The latency results (clock cycles) of the \textit{reference} execution, are listed in the fourth to the sixth column. The latency results of the \textit{hybrid} execution  are shown in the seventh column to the the ninth column. \enquote{Min} and \enquote{Max} mean  the minimum latency and the maximum latency of a sample in the inference dataset executing on the RISC-V CPU, respectively. \enquote{Avg} indicates the average latency of all the samples. 
Considering the relative latency comparison of the reference and hybrid execution, both the minimum latency and the average latency can be reduced effectively using logic flows. For example, the minimum latency can be reduced by up to \SI{52.2}{\percent}. 
%The latency reduction of all the neural networks using hybrid execution in \textit{NN2Logic} is shown in the tenth column to the twelfth column. According to these results, both the minimum latency and the average latency with \textit{NN2Logic} can be reduced effectively. 
%These reduction comes from the reduced number of MAC operations and taking advantages of the logic circuit in CPUs. 
For the maximum latency, there is a slight increase due to additional checks when executing the sample  using the original form of the neural network.
%To clearly compare the latency of all the neural networks of hybrid code with \textit{NN2Logic}, we visualize the latency distributions with hybrid code and reference code, as shown in \figname~\ref{ibex-occ-ii}. 
%The hybrid execution can mainly reduce the latency of layers two and three. %, since with the equivalent decision tree of the neural network, we mostly find logic redundancy between these layers and the first layer.
The latency reduction considering only the second and the third layers can be up to \SI{39.3}{\percent}.
%The latency reduction considering only these layers are listed in the thirteenth column in Table \ref{tab:first-table}. According to this column, the latency can be reduced by up to \SI{39.3}{\percent} in those layers. 
%
Finally, the last column in \cref{tab:first-table} demonstrates the percentage of samples in the inference dataset exiting through logic flows of the neural network. %This column shows that \textit{NN2Logic} can benefit the execution of many samples in the inference dataset on CPUs.

\section{Conclusion}\label{sec:conclusion}
In this paper, we have proposed to convert the execution of neural networks into logic flows for edge computing.
It opens up a new dimension to accelerate neural networks on edge devices with CPUs. %It can potentially shed light on the logic redundancy inside neural networks. 
Future work includes extending the framework into deep neural networks with multi-class classification.
%The framework itself can be further improved by using a solver that returns all the IIS of an infeasible problem and by choosing a better path selection approach.

% Bibliography
\clearpage
\enlargethispage{-10cm}
\printbibliography{}

\end{document}